\documentclass[conference]{IEEEtran}
\IEEEoverridecommandlockouts

\usepackage{cite}
\usepackage{amsmath,amssymb,amsfonts}
\usepackage{algorithm}
\usepackage{algorithmic}
\usepackage{graphicx}
\usepackage{textcomp}
\usepackage{xcolor}

\usepackage{booktabs} 
\usepackage{hyperref}

\def\BibTeX{{\rm B\kern-.05em{\sc i\kern-.025em b}\kern-.08em
    T\kern-.1667em\lower.7ex\hbox{E}\kern-.125emX}}

\begin{document}

\title{From Theory to Throughput: CUDA-Optimized APML for Large-Batch 3D Learning
\thanks{This research was supported by the Research Council of Finland 6G Flagship Programme (Grant 346208), Horizon Europe CONVERGE (Grant 101094831), and Business Finland WISECOM (Grant 3630/31/2024).}
}

\author{%
\IEEEauthorblockN{Sasan Sharifipour\textsuperscript{*}, Constantino \'Alvarez Casado\textsuperscript{*}, Manuel Lage Ca\~nellas, Miguel Bordallo L\'opez}
\IEEEauthorblockA{\textit{Center for Machine Vision and Signal Analysis (CMVS), University of Oulu}, Finland}
\thanks{\textsuperscript{*}These authors contributed equally to this work.}
}

\maketitle

\begin{abstract}

Loss functions are fundamental to learning accurate 3D point cloud models, yet common choices trade geometric fidelity for computational cost. Chamfer Distance is efficient but permits many-to-one correspondences, while Earth Mover Distance better reflects one-to-one transport at high computational cost. APML approximates transport with differentiable Sinkhorn iterations and an analytically derived temperature, but its dense formulation scales quadratically in memory. We present CUDA-APML, a sparse GPU implementation that thresholds negligible assignments and runs adaptive softmax, bidirectional symmetrization, and Sinkhorn normalization directly in COO form. This yields near-linear memory scaling and preserves gradients on the stored support, while pairwise distance evaluation remains quadratic in the current implementation. On ShapeNet and MM-Fi, CUDA-APML matches dense APML within a small tolerance while reducing peak GPU memory by 99.9\%. Code available at: \url{https://github.com/Multimodal-Sensing-Lab/apml}


\end{abstract}


\section{Introduction}

Three-dimensional point clouds represent geometry from LiDAR, depth cameras, and structured-light scanners, and they also appear in cross-modal pipelines that infer 3D structure from non-visual measurements. Point cloud learning supports robotics, AR and VR, wireless human sensing, and digital twin pipelines \cite{Guo2020Survey,Qi2017PointNet,Maetta2025CSI2PC,xue2020lidar}. In supervised settings, models output unordered point sets, so training requires permutation-invariant losses that remain stable at large point counts.

Matching quality and scalability are often in tension. Nearest-neighbor objectives are efficient but allow many-to-one assignments that can degrade geometric coverage, whereas optimal transport losses better reflect one-to-one matching but are costly at training scale. This cost constrains batch size and point count, reduces the number of feasible training runs under fixed compute budgets, and increases energy and monetary cost during GPU training and tuning. APML \cite{Sharifipour2025APML} narrows this gap by constructing a differentiable transport plan with temperature-scaled similarities and Sinkhorn normalization, using an analytically derived temperature. However, its dense formulation stores pairwise matrices with quadratic memory growth, which can dominate training resources. For example, \cite{Sharifipour2025APML} reported close to 300~GB of GPU memory when training FoldingNet \cite{Yang2018FoldingNet} on ShapeNet-55 \cite{Chang2015ShapeNet}. Such requirements restrict use in higher-resolution regimes that arise in large scenes and repeated updates, including mapping and digital twin pipelines \cite{dai2017scannet,behley2019semantickitti,caesar2020nuscenes,hackel2017semantic3d,landrieu2018large,wysocki2025tum2twin}.

We present CUDA-APML, a sparse GPU implementation of APML that exploits the empirical sparsity of the transport plan after adaptive softmax thresholding. CUDA-APML constructs bidirectional assignments directly in COO form, performs on-device symmetrization and Sinkhorn normalization on the sparse support, and computes the loss and gradients without materializing dense matrices. On ShapeNet and MM-Fi, CUDA-APML matches dense APML within a small tolerance while reducing peak GPU memory by up to 99.9\%, improving the practical cost profile of transport-based supervision for point cloud and cross-modal 3D learning.

\section{Related Work}
Point cloud learning requires permutation-invariant losses that compare unordered predicted point sets with reference point sets. Chamfer Distance (CD) is widely used because it reduces the comparison to nearest-neighbor distances computed in both directions \cite{Fan2017PCN,Qi2017PointNet}. However, its many-to-one assignments can cause point clustering in dense regions and weak coverage in sparse regions \cite{Lin2023HyperCD}. Extensions such as density-aware CD (DCD) \cite{Wang2021DCD}, HyperCD \cite{Lin2023HyperCD}, and InfoCD \cite{Lin2023InfoCD} mitigate some of these effects but still rely on nearest-neighbor matching and remain sensitive to sampling density and local ambiguities. By contrast, Earth Mover Distance (EMD) enforces one-to-one transport and better reflects structural alignment \cite{Rubner2000EMD}, but its computational cost and common equal-cardinality constraints limit its use in training. Alternatives include projected distances such as sliced Wasserstein and approximate transport solvers \cite{Wu2019SWD,Genevay2018Sinkhorn}, which often introduce additional design choices and may not match full transport behavior on heterogeneous point distributions.

Entropy-regularized optimal transport provides a differentiable approximation of EMD that can be computed by Sinkhorn normalization \cite{Cuturi2013Sinkhorn,Peyre2019OT}. Sinkhorn-based losses have been used for soft correspondence problems such as registration and feature matching \cite{Sarlin2020SuperGlue,Shen2021RobustOT}, but typically require tuning the regularization strength to balance sharpness and stability. APML \cite{Sharifipour2025APML} builds a temperature-scaled similarity matrix from pairwise distances and applies Sinkhorn normalization to obtain soft correspondences with approximately one-to-one marginals. Its temperature is derived from local distance gaps to enforce a minimum assignment probability, reducing manual tuning, but its dense formulation stores and updates pairwise matrices with quadratic memory growth, which restricts batch size and point count in practice \cite{Sharifipour2025APML}.

Scaling optimal transport has been studied through sparse or structured support restriction and low-rank approximations \cite{Blondel2018SparseOT,Li2023ImportanceSparsification,Tang2024SparseNewton}. In parallel, kernel-level designs show that removing dense intermediates can shift practical limits for quadratic operators, as in FlashAttention \cite{dao2022flashattention} and KeOps \cite{charlier2021kernel}. In 3D perception, sparse tensor engines and point-based libraries provide GPU support for irregular sparsity patterns to improve memory behavior and throughput \cite{choy20194d,tang2022torchsparse}. These directions motivate a sparse and GPU-oriented implementation of APML. In this work, we exploit the empirical sparsity of the APML transport plan after adaptive softmax thresholding and run symmetrization and Sinkhorn normalization on the sparse support to reduce the dense matrix footprint while keeping the objective within a controlled approximation.

\section{Proposed Methodology}
\label{sec:method}

This section presents CUDA-APML, a sparse GPU implementation of APML \cite{Sharifipour2025APML}. The objective is to avoid dense $N\times M$ tensors during forward and backward passes while keeping the APML formulation and its differentiability in the sparse support. CUDA-APML constructs row-wise and column-wise adaptive-softmax assignments directly in sparse coordinate (COO) form, symmetrizes them on device, applies Sinkhorn normalization on the sparse support, and evaluates the loss using only stored nonzero entries.

\subsection{Background: dense APML}
\label{sec:apml_background}

Let $X=\{x_i\}_{i=1}^{N}$ be the predicted points and $Y=\{y_j\}_{j=1}^{M}$ be the reference points with $x_i,y_j\in\mathbb{R}^{d}$. APML starts from the pairwise cost matrix $C\in\mathbb{R}^{N\times M}$ with \( C_{ij}=\lVert x_i-y_j\rVert_2 \). For a generic cost vector $c\in\mathbb{R}^{K}$, APML shifts it as $\tilde c_k=c_k-\min_{\ell}\,c_\ell$ so that $\min_k \tilde c_k=0$. Let $\tilde c^{(2)}$ denote the second smallest value in $\tilde c$. The local gap is defined as $g=\tilde c^{(2)}+\delta$ with $\delta>0$. APML computes a temperature $T$ that enforces a minimum probability mass $p_{\min}$ on the smallest entry, under the approximation in \cite{Sharifipour2025APML}, given by:
\begin{equation}
T=\frac{-\log\!\left(\frac{1-p_{\min}}{(K-1)p_{\min}}\right)}{g},
\qquad 0<p_{\min}<1,\quad K>1.
\label{eq:temp_adapt}
\end{equation}
When $\tilde c^{(2)}<\epsilon_g$, APML applies a uniform fallback distribution for numerical stability.

APML applies the adaptive-temperature softmax row-wise and column-wise on $C$, producing $P_{\text{row}}$ and $P_{\text{col}}$, and initializes \(P^{(0)}=\frac{1}{2}(P_{\text{row}}+P_{\text{col}})\). Sinkhorn normalization then alternates column and row scaling for $L_{\text{iter}}$ iterations with stability constant $\epsilon_{\text{stab}}$ \cite{Sharifipour2025APML}. The APML objective is the Frobenius inner product
\(\mathcal{L}_{\text{APML}}=\langle P,C\rangle_F=\sum_{i=1}^{N}\sum_{j=1}^{M}P_{ij}C_{ij}.\)
The dense formulation stores and updates $C$ and $P$, which yields $O(NM)$ memory.

\subsection{Sparsity of the APML initialization}
\label{sec:sparsity_characterization}

CUDA-APML is based on an empirical property reported in \cite{Sharifipour2025APML} and confirmed in our measurements. After the adaptive softmax stage, most entries in the unnormalized similarity matrix are close to zero. For a fixed row $i$, the unnormalized similarity can be written as \(s^{\text{row}}_{ij}=\exp\!\left(-T_i\,(C_{ij}-C_{i\min})\right),\) with \(C_{i\min}=\min_{j} C_{ij},\) where $T_i$ is computed from the row-wise gap through Eq.~\eqref{eq:temp_adapt}. The corresponding normalized probabilities are defined as:
\vspace{-3mm}
\begin{equation}
p^{\text{row}}_{ij}
=
\frac{s^{\text{row}}_{ij}}{\sum_{k=1}^{M}s^{\text{row}}_{ik}}.
\label{eq:row_prob}
\end{equation}
Analogous definitions hold for the column-wise direction, producing $p^{\text{col}}_{ij}$.

CUDA-APML introduces a pruning threshold $\tau$ applied to the unnormalized similarities and keeps only the index pairs in the support sets, defined as \(\Omega^{\text{row}}_{\tau} = \left\{(i,j)\,:\, s^{\text{row}}_{ij}\ge \tau\right\},\) and \(\Omega^{\text{col}}_{\tau}=\left\{(i,j)\,:\, s^{\text{col}}_{ij}\ge \tau\right\}.\) The sparse initialization keeps the union support $\Omega_{\tau}=\Omega^{\text{row}}_{\tau}\cup\Omega^{\text{col}}_{\tau}$ and stores only those entries in COO format. The objective is to reduce memory traffic and storage while preserving the probability mass that dominates the transport plan.

\subsection{Sparse COO adaptive softmax and Sinkhorn on GPU}
\label{sec:sparse_ops}

The number of kept entries per row and per column is data-dependent, so CUDA-APML constructs COO buffers in two GPU passes per direction. In the first pass, each thread block scans one row or column, computes costs $C_{ij}$ on the fly, tracks the minimum and second minimum to obtain the local temperature, and counts entries that satisfy the pruning rule in $\Omega^{\text{row}}_{\tau}$ or $\Omega^{\text{col}}_{\tau}$. An exclusive prefix sum converts these counts into write offsets and yields the total number of nonzeros. In the second pass, the kernel rescans, recomputes unnormalized similarities (with $C_{i\min}$ subtracted before the exponential), writes the kept COO triples $(\mathbf{i},\mathbf{j},\mathbf{v})$, and normalizes within each row or column by the sum of kept similarities on that support. If the uniform fallback is triggered, the kernel writes a uniform distribution over the corresponding row or column. This produces $P_{\text{row}}$ on $\Omega^{\text{row}}_{\tau}$ and $P_{\text{col}}$ on $\Omega^{\text{col}}_{\tau}$.

To symmetrize without dense materialization, CUDA-APML concatenates the COO streams, maps each $(i,j)$ to a 64-bit key (for example $\texttt{key}=i\cdot M + j$), sorts by key on device, and reduces duplicates by averaging values, yielding a single COO plan $P^{(0)}$ on $\Omega_{\tau}$. Sinkhorn normalization then operates on this sparse support. Let $P$ be stored in COO with entries $\{(i_t,j_t,v_t)\}_{t=1}^{\text{nnz}}$. One iteration alternates column scaling,
\begin{equation}
v^{(\ell+\frac{1}{2})}_{t}
=
\frac{v^{(\ell)}_{t}}{\sum_{t':\, j_{t'}=j_t} v^{(\ell)}_{t'} + \epsilon_{\text{stab}}},
\label{eq:coo_col_norm}
\end{equation}
and row scaling,
\begin{equation}
v^{(\ell+1)}_{t}
=
\frac{v^{(\ell+\frac{1}{2})}_{t}}{\sum_{t':\, i_{t'}=i_t} v^{(\ell+\frac{1}{2})}_{t'} + \epsilon_{\text{stab}}},
\label{eq:coo_row_norm}
\end{equation}
implemented by an accumulation kernel (atomic adds into dense length-$M$ or length-$N$ buffers) followed by a scaling kernel that updates all COO values. This keeps the plan sparse and avoids any $N\times M$ tensor. We use $L_{\text{iter}}$ as in dense APML \cite{Sharifipour2025APML}.

\subsection{Loss, gradients, and stability}
\label{sec:sparse_loss}

After Sinkhorn normalization, CUDA-APML evaluates the objective on the COO support as
\(\mathcal{L}_{\text{CUDA-APML}}=\sum_{t=1}^{\text{nnz}} v_t \,\lVert x_{i_t}-y_{j_t}\rVert_2.\)
No dense cost matrix $C$ is formed, and the forward pass computes distances only for the stored index pairs. The backward pass differentiates through the same sparse computation graph. In particular, the distance term contributes the standard pairwise gradient on the stored pairs:
\begin{equation}
\frac{\partial}{\partial x_{i_t}} \lVert x_{i_t}-y_{j_t}\rVert_2
=
\frac{x_{i_t}-y_{j_t}}{\lVert x_{i_t}-y_{j_t}\rVert_2+\epsilon_{\text{dist}}},
\label{eq:dist_grad}
\end{equation}
where $\epsilon_{\text{dist}}>0$ avoids division by zero when two points coincide. Gradients of the transport weights $v_t$ with respect to $X$ and $Y$ are computed through the fused kernels for the adaptive softmax and Sinkhorn updates on the sparse support. Pruned entries are treated as zero and do not contribute to forward or backward passes, which suppresses weak competing correspondences and sharpens the dominant gradient signal.

CUDA-APML follows the stability conditions of APML \cite{Sharifipour2025APML}. Stability is ensured by clamping the local gap as $\max(\text{gap}, \epsilon_g)$ instead of uniform fallback. Sinkhorn normalization uses $\epsilon_{\text{stab}}$, and sparsity is controlled by the pruning threshold $\tau$.

\subsection{Complexity, memory, and implementation details}
\label{sec:complexity_impl}

Algorithm~\ref{alg:sparse_apml} summarizes the CUDA-APML pipeline for one batch and highlights where dense tensors are avoided.

\begin{algorithm}[ht!]
\caption{CUDA-APML for one batch}
\label{alg:sparse_apml}
\begin{algorithmic}[1]
\REQUIRE Predicted $X\in\mathbb{R}^{B\times N\times d}$, reference $Y\in\mathbb{R}^{B\times M\times d}$, $p_{\min}$, $\tau$, $L_{\text{iter}}$
\STATE \textbf{Row direction:} scan each row, compute minima and temperature, count pairs with similarity $\ge\tau$, exclusive scan to obtain offsets, rescan to write and normalize COO
\STATE \textbf{Column direction:} repeat with swapped roles to obtain $P_{\text{col}}$
\STATE \textbf{Symmetrize:} concatenate COO, sort by 64-bit key $(i,j)$, reduce duplicates by averaging to obtain $P^{(0)}$
\FOR{$\ell=1$ to $L_{\text{iter}}$}
  \STATE Column scaling in COO using dense column-sum buffer and $\epsilon_{\text{stab}}$
  \STATE Row scaling in COO using dense row-sum buffer and $\epsilon_{\text{stab}}$
\ENDFOR
\STATE \textbf{Loss and backward:} evaluate loss on COO pairs and backpropagate through sparse operators
\end{algorithmic}
\end{algorithm}


Let $B$ be the batch size, let $d$ be the point dimension, and let $\text{nnz}=|\Omega_{\tau}|$ be the number of stored COO entries after symmetrization. Dense APML stores the cost matrix $C$ and transport plan $P$ as $B\times N\times M$ tensors, which yields $O(BNM)$ memory. With single precision, storing both requires at least $2\cdot 4\cdot BNM$ bytes, excluding intermediate buffers from the row-wise and column-wise stages. The arithmetic cost is dominated by pairwise distance evaluation and dense Sinkhorn updates, scaling as $O(BNM d + BNM L_{\text{iter}})$. CUDA-APML avoids materializing $C$ and $P$ as dense tensors and stores the plan in COO. With 32-bit indices $(\mathbf{i},\mathbf{j})$ and 32-bit values $\mathbf{v}$, the plan storage is $12\,\text{nnz}$ bytes, plus temporary buffers for symmetrization and Sinkhorn. Symmetrization adds a 64-bit key per entry and sorting workspace. Sinkhorn uses two dense 1D accumulation buffers of lengths $N$ and $M$ for row and column sums. Overall, memory scales as \(O\!\left(B(\text{nnz}+N+M)\right)\) up to constant factors from sorting and scan workspaces.

In the current implementation, the adaptive softmax stage still scans all pairs to identify minima and apply thresholding, so distance evaluation remains $O(BNM d)$. The runtime reduction comes from lower memory traffic and from restricting Sinkhorn updates and loss evaluation to $\text{nnz}$ entries. Sinkhorn iterations cost $O(B\,\text{nnz}\,L_{\text{iter}})$, and symmetrization is dominated by sorting, with $O(\text{nnz}\log \text{nnz})$ in the comparison model. Unless stated otherwise, we use $\tau=10^{-8}$, $L_{\text{iter}}=10$, and $\epsilon_{\text{stab}}=10^{-8}$, and we set $p_{\min}$ as in \cite{Sharifipour2025APML}.

\section{Evaluation Methodology}
\label{sec:evaluation}

\subsection{Datasets, models, and baselines}
We evaluate CUDA-APML on point cloud completion and cross-modal generation. For completion, we use ShapeNet-34 and ShapeNet-55 \cite{Chang2015ShapeNet} with the standard partial-to-complete pairs and official splits, and PCN \cite{Yuan2018PCN} (2,048-point partial input and 16,384-point reference). For cross-modal generation, we use MM-Fi \cite{Yang2024MMFI,Maetta2025CSI2PC}, which pairs WiFi CSI with LiDAR-based 3D human point clouds, and we follow the CSI2PC \texttt{manual\_split} protocol that tests generalization to unseen subjects and environments \cite{Maetta2025CSI2PC}. We integrate losses into FoldingNet \cite{Yang2018FoldingNet}, PCN \cite{Yuan2018PCN}, and PoinTr \cite{Yu2021PoinTr} for completion, and into CSI2PC \cite{Maetta2025CSI2PC} for MM-Fi. We compare CUDA-APML against CD \cite{Fan2017PCN}, DCD \cite{Wang2021DCD}, HyperCD \cite{Lin2023HyperCD}, InfoCD \cite{Lin2023InfoCD}, and dense APML \cite{Sharifipour2025APML}.

\begin{figure*}[ht!]
  \centering
  \includegraphics[width=\linewidth]{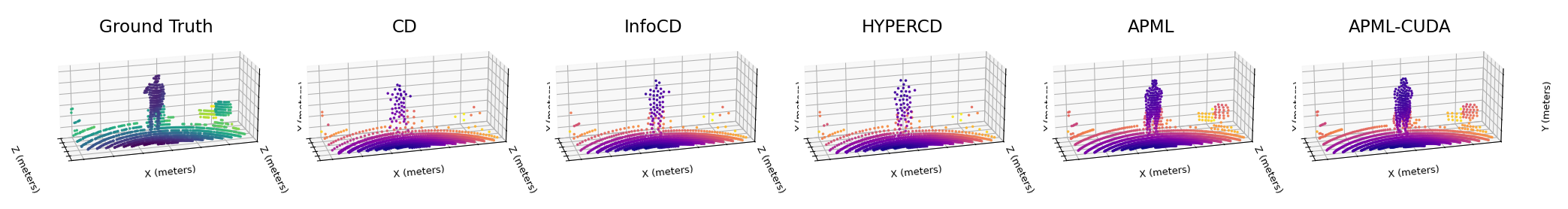}
  \caption{Qualitative comparison on an MM-Fi sample. From left to right: ground truth and predictions obtained with CD, InfoCD, HyperCD, APML, and CUDA-APML.}
  \label{fig:qualitative}
  \vspace{-4mm}
\end{figure*}

\subsection{Metrics and protocol}
\label{subsec:metrics_protocol}
We report CD in $\ell_1$ and $\ell_2$ variants \cite{Fan2017PCN}, EMD$\times 100$ \cite{Rubner2000EMD}, and F1-score at threshold $0.01$ \cite{Yuan2018PCN}. Efficiency is measured by wall-clock time per epoch and peak GPU memory. Each configuration is trained three times with different seeds and we report the mean. ShapeNet results use the standard test split \cite{Chang2015ShapeNet} and the default training protocols of each backbone, while MM-Fi uses \texttt{manual\_split} \cite{Maetta2025CSI2PC}. \textcolor{black}{To support the sparse formulation in Sec.~\ref{sec:method}, we also report (i) the number of nonzero COO entries after symmetrization as a function of point count on synthetic random point sets using $500$ trials per point count, and (ii) peak memory per sample for dense APML and CUDA-APML in the same synthetic setup.}


\subsection{Implementation details and measurement}
\label{subsec:impl_measurement}
All experiments use PyTorch 2.5 with CUDA 12.8 and are run on NVIDIA V100 (32\,GB) and RTX 4070 GPUs. Unless stated otherwise, we use Adam with learning rate $10^{-4}$ and batch size $32$, and we follow the official training schedules and preprocessing of each backbone. For CUDA-APML, COO construction runs on device, Sinkhorn uses $L_{\text{iter}}=10$ and $\epsilon_{\text{stab}}=10^{-8}$, and the pruning threshold is $\tau=10^{-8}$ unless stated otherwise. Runtime per epoch is measured with the PyTorch profiler, and peak GPU memory is measured using PyTorch CUDA memory statistics and NVIDIA tools, including \texttt{nvidia-smi} and CUDA profiler interfaces, reporting the maximum observed value per run.


\section{Experimental Results}
\label{sec:results}

\subsection{Quantitative results on ShapeNet and MM-Fi}
Table~\ref{tab:results} reports the main benchmarks (mean of three runs). On ShapeNet-34 with FoldingNet, the similar results for both Dense-APML and CUDA-APML improve EMD over CD, HyperCD, and InfoCD and achieve the highest F1-score. On MM-Fi with CSI2PC, CUDA-APML improves EMD over DCD with slightly better results than Dense-APML.

\vspace{-3mm}
\begin{table}[ht!]
\def\arraystretch{1.05}
\setlength{\tabcolsep}{0.5em}
  \centering
  \caption{Comparison of loss functions on ShapeNet-34 (completion) and MM-Fi (generation).
  Metrics: F1-score (↑) and EMD$\times 100$ (↓).}
  \label{tab:results}
  \begin{tabular}{lccc}
    \toprule
    Dataset / Model & Loss & F1-score ↑ & EMD$\times 100$ ↓ \\
    \midrule
    ShapeNet-34 / FoldingNet & CD & 0.11 & 27.7 \\
                             & HyperCD & 0.19 & 16.5 \\
                             & InfoCD & 0.18 & 22.3 \\
                             & Dense-APML & \textbf{0.20} & \textbf{9.5} \\
                             & CUDA-APML & \textbf{0.20} & \textbf{9.5} \\
    \midrule
    MM-Fi / CSI2PC           & DCD & -- & 25.7 \\
                             & Dense APML & -- & 16.3 \\
                             & CUDA-APML & -- & \textbf{16.1} \\
    \bottomrule
  \end{tabular}
\end{table}

\subsection{Efficiency and scaling behavior}
\begin{figure}[ht!]
  \centering
  \includegraphics[width=\linewidth]{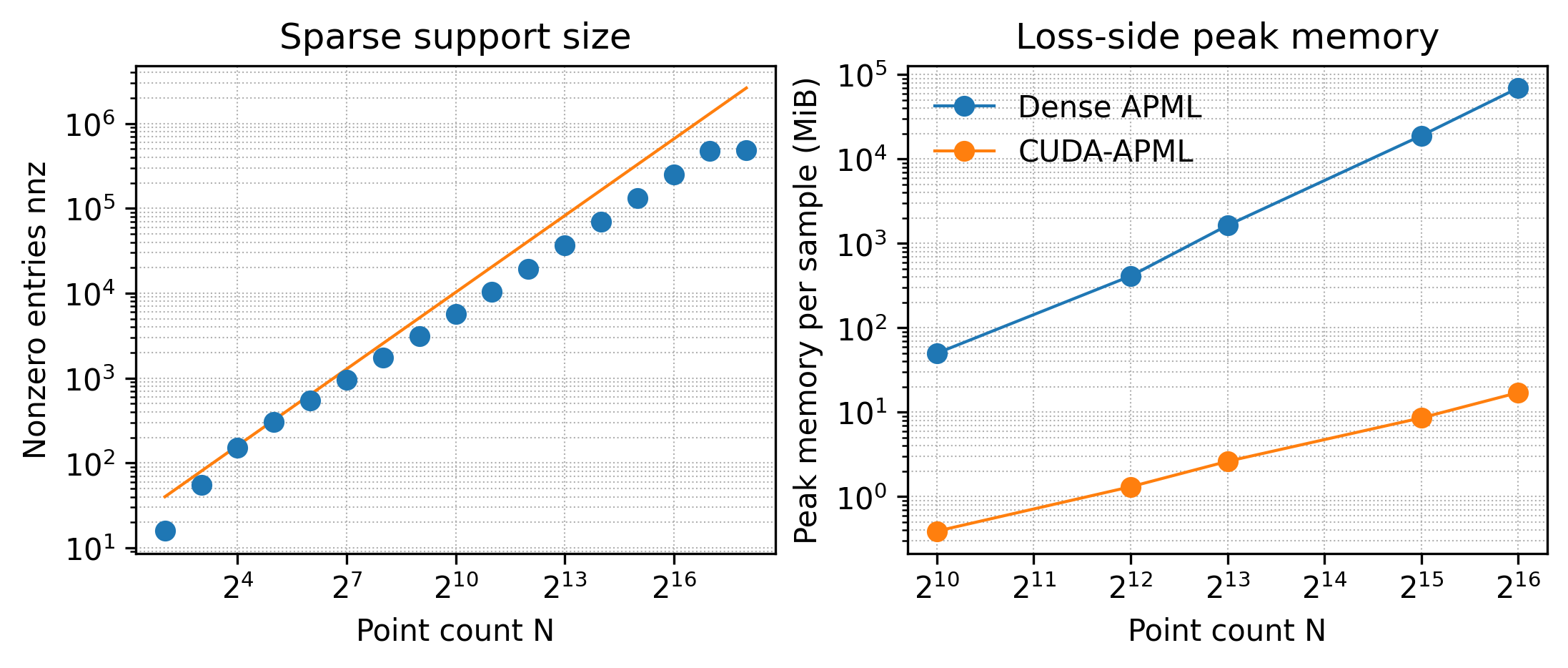}
  \vspace{-6mm}
  \caption{Scaling measurements for CUDA-APML on synthetic point sets. Left: number of nonzero COO entries after symmetrization as a function of point count (500 trials per $N$). Right: peak memory per sample for dense APML and CUDA-APML in the same setup.}
  \label{fig:scaling}
\end{figure}

Table~\ref{tab:efficiency} shows peak GPU memory for FoldingNet on ShapeNet-55 (batch size 32). CUDA-APML reduces memory across all tested point counts. Figure~\ref{fig:scaling} reports synthetic scaling: the left panel shows the number of COO nonzeros after symmetrization (500 trials per $N$ up to $262{,}144$), and the right panel shows peak loss-side memory per sample for dense APML and CUDA-APML in the same setup. Runtime per epoch and peak GPU memory are measured as described in Sec.~\ref{subsec:impl_measurement}.

\begin{table}[ht!]
\def\arraystretch{1.05}
\setlength{\tabcolsep}{2.2em}
  \centering
  \caption{Peak GPU memory for FoldingNet on ShapeNet-55.}
  \label{tab:efficiency}
  \begin{tabular}{lcc}
    \toprule
    Point count & Dense APML & CUDA-APML \\
    \midrule
    1k  & 50 MB  & 0.39 MB \\
    8k  & 1.6 GB & 2.6 MB \\
    32k & 18.5 GB & 8.6 MB \\
    65k & 68 GB  & 17.1 MB \\
    \bottomrule
  \end{tabular}
\end{table}

\subsection{Qualitative results}
Figure~\ref{fig:qualitative} compares ground truth point clouds with predictions from CD, InfoCD, HyperCD, APML, and CUDA-APML. In the shown MM-Fi sample, Chamfer-based losses yield more isolated points and gaps in the human silhouette. APML and CUDA-APML produce more coherent coverage with fewer missing regions. 



\section{Conclusion}

We presented CUDA-APML, a sparse CUDA implementation of APML that removes the quadratic memory bottleneck of dense transport plan construction for point cloud supervision. CUDA-APML constructs row-wise and column-wise adaptive-softmax assignments in COO form, performs on-device symmetrization, and applies Sinkhorn normalization on the sparse support without materializing dense $N\times M$ tensors. Across ShapeNet and MM-Fi, CUDA-APML preserves the accuracy trends of dense APML while reducing peak GPU memory by more than 99.9\% and showing near-linear scaling of the stored support and loss-side memory in synthetic tests, which makes APML practical when dense transport losses are memory limited.

\textcolor{black}{A limitation is that the adaptive-softmax stage still scans all pairwise distances, so distance evaluation remains quadratic in the number of points. Future work will restrict the candidate support, for example via neighborhood pruning or approximate nearest neighbors, and will evaluate the method on larger scene-level and digital twin update settings.}

\section*{Acknowledgment}
The research was supported by the Research Council of Finland (6G Flagship, Grant 346208), Horizon Europe CONVERGE (Grant 101094831), and Business Finland WISECOM (Grant 3630/31/2024). Sasan Sharifipour acknowledges the funding from the Finnish Doctoral Program Network on Artificial Intelligence, AI-DOC (decision number VN/3137/2024-OKM-6), supported by Finland’s Ministry of Education and Culture and hosted by the Finnish Center for Artificial Intelligence (FCAI).The authors acknowledge CSC–IT Center for Science, Finland, for computational resources.

\bibliographystyle{IEEEtran}
\bibliography{references}

\end{document}